\documentclass{article} % For LaTeX2e
\usepackage[preprint]{neurips_2023}

\usepackage{amsmath,amsfonts,bm}

\def\eqref#1{equation~\ref{#1}}
\def\1{\bm{1}}

\DeclareMathAlphabet{\mathsfit}{\encodingdefault}{\sfdefault}{m}{sl}
\SetMathAlphabet{\mathsfit}{bold}{\encodingdefault}{\sfdefault}{bx}{n}

\usepackage{wrapfig}
\usepackage{tikz}
\usepackage{subfigure}
\usetikzlibrary{positioning}
\usetikzlibrary{arrows.meta}
\usepackage{multirow}
\usepackage{booktabs}

\usepackage{makecell}
\usepackage{adjustbox}
\usepackage{enumitem}
\usepackage{hyperref}
\usepackage{url}
\usepackage{algorithm}
\usepackage{algorithmic}

\usepackage{amsthm}
\newtheorem{theorem}{Theorem}[section]
\newtheorem{proposition}[theorem]{Proposition}

\theoremstyle{definition}
\newtheorem{definition}[theorem]{Definition}

\theoremstyle{remark}

\title{Defining and Measuring Feature Diversity through the Lens of Category Theory}

\title{Going Beyond Neural Network Feature Similarity: Network Feature Complexity and An Interpretation through the Lens of Category Theory}
\title{Going Beyond Neural Network Feature Similarity: The Network Feature Complexity and Its Interpretation Using Category Theory}
\author{Yiting Chen, Zhanpeng Zhou \& Junchi Yan \thanks{Corresponding author. This work was partly supported by National Key Research and Development Program of China (2020AAA0107600), NSFC (62222607) and STCSM (22511105100).} \\
Department of Computer Science and Engineering and MoE Key Lab of Artificial Intelligence\\
Shanghai Jiao Tong University\\
\texttt{\{sjtucyt,zzp1012,yanjunchi\}@sjtu.edu.cn}
}

\begin{document}

%are often regarded as ``black box" models due to the lack of comprehensive understanding of their inner mechanism and behavior. 
\maketitle

%\vspace{-10pt}
\begin{abstract}
The behavior of neural networks still remains opaque, and a recently widely noted phenomenon is that networks often achieve similar performance when initialized with different random parameters. This phenomenon has attracted significant attention in measuring the similarity between features learned by distinct networks. However, feature similarity could be vague in describing the same feature since equivalent features hardly exist. In this paper, we expand the concept of equivalent feature and provide the definition of what we call \emph{functionally equivalent features}. These features produce equivalent output under certain transformations. 
Using this definition, we aim to derive a more intrinsic metric for the so-called \emph{feature complexity} regarding the redundancy of features learned by a neural network at each layer. We offer a formal interpretation of our approach through the lens of category theory, a well-developed area in mathematics. To quantify the feature complexity, we further propose an efficient algorithm named Iterative Feature Merging. Our experimental results validate our ideas and theories from various perspectives. We empirically demonstrate that the functionally equivalence widely exists among different features learned by the same neural network and we could reduce the number of parameters of the network without affecting the performance. We have also drawn several interesting empirical findings, including: 
1) the larger the network, the more redundant features it learns; 2) in particular, we show how to prune the networks based on our finding using direct equivalent feature merging, without fine-tuning which is often needed in peer network pruning methods; 3) same structured networks with higher feature complexity achieve better performance; 4) through the layers of a neural network, the feature complexity first increase then decrease; 5) for the image classification task, a group of functionally equivalent features may correspond to a specific low-level semantic meaning.
Source code will be made publicly available.
\end{abstract}

%\vspace{-10pt}
\section{Introduction}
%\vspace{-5pt}
Deep neural networks (DNNs) have achieved significant success across diverse fields, including vision, texts among other areas. However, DNNs are often regarded as ``black box" models associated with high dimensional feature maps and numerous parameters for training.
Recently, many studies report an interesting phenomenon that neural networks with different random initialization often converge to solutions with similar performance on the test set~\citep{Dauphin14nips,frankle2020linear}. Meanwhile, methods such as pruning~\citep{Wen16nips,Ye18iclr,PengWCH19icml} and knowledge distillation~\citep{Hinton15} have been proposed to reduce the complexity of the neural network structure, leading to a more compact DNN achieving similar performance to their dense/large counterparts.
These phenomena further draw increasing studies on the features\footnote{Precisely speaking, note that the term \emph{feature} used in this paper has two different meanings that can be (hopefully) distinguished from the context: 1) in its specific meaning, the layer-wise feature map corresponding to a data point as input to the trained network; 2) more generally, the characteristic or measurable properties of the (training) data that are learned by the neural network, which is in fact an function of training data, training algorithm and model structure. In fact, most previous works, \emph{e.g.} measuring feature similarity~\citep{li2015convergent,RaghuGYS17nips, Wang18nips,Williams21nips, Barannikov22ICML}, are based on a set of feature maps corresponding to a specific dataset (one layer-wise feature map for one input sample). Our definition and method, as will be shown later, correspond more to the general meaning of feature, which are input-agnostic and thus perhaps more reflects the nature of the network.} learned by neural networks and a popular treatment is measuring the so-called feature similarity\footnote{The feature similarity is also referred to as representational similarity.}. Various feature similarity measures~\citep{li2015convergent,Morcos18nips} have been devised to quantify the distance between two features, and some works have further empirically found that similar features can be learned from either between different networks~\citep{li2015convergent,Wang18nips} or within a (wide) network~\citep{Nguyen21iclr}. In literature, feature similarity (or sometimes distance) measures are often designed to be invariant to certain transformations under certain transformations \emph{e.g.} permutation~\citep{li2015convergent}, isotropic scaling~\citep{Barannikov22ICML}, invertible linear transformation~\citep{Wang18nips}, affine transformation~\citep{RaghuGYS17nips,Williams21nips} etc. 
However, the exact ``equivalent features" in the sense of numerically zero distance can hardly exist in practice. Therefore, previous works are often vague in exactly describing the inherent structure of the neural networks (\emph{e.g.} the redundancy of features learned by a neural network or the same feature learned by different neural networks) and can hardly provide actionable insights \emph{e.g.} which features can be merged (when they are equivalent features).

In this paper, we expand the concept of feature equivalence in feature similarity literature~\citep{li2015convergent,RaghuGYS17nips, Wang18nips,Williams21nips, Barannikov22ICML} and formally define a more general form: the so-called functional equivalence. Features are considered functionally equivalent if they produce outputs that are equivalent under certain transformations. Using the concept of functionally equivalent features, one can derive a metric to reflect the inherent feature complexity of the neural network regarding the redundancy of features learned by neural networks. In another words, the feature complexity corresponds to the most compact representation of a neural network that one can get.

For a clearer and more formal description of our methodology, we provide an abstract view of neural networks and interpret our approach using the language of category theory, a branch of mathematics, which has also been applied in different fields from physics to computer science. In general, a category is a graph with objects and arrows (or ``morphisms") between objects. The mappings between categories are called functors that preserves the structure and the mappings between functors are called natural transformations (for a formal introduction, see Section~\ref{sec:CT}). In the context of neural networks, we abstractly represent the network structure as a category and a certain neural network as a functor that maps this structure to specific parameters.
Through the lens of category theory, we show that functional equivalence between features can be elegantly represented as the existence of a natural transformation between two functors defined above. While functional equivalence is non-trivial, we specifically observe that the extensively explored empirical phenomenon known as linear mode connectivity (LMC)~\citep{frankle2020linear} actually indicates functional equivalence between features. With the defined functionally equivalent features, we further consider the feature complexity of a neural network regarding the natural transformations between a functor and itself, where the existence of more than one natural transformation indicates the redundancy in features.

To empirically measure the feature complexity, we propose an algorithm called Iterative Feature Merging (IFM) to merge the features that are functionally equivalent. By iteratively matching the weights corresponding to different features, we merge functionally equivalent features. Experimental results show that DNNs learns a lot of functionally equivalent features which could be merged or removed with little impact on the neural network performance. We also provide many valuable insights that may inspire future works. Our major contributions are as follows:

\begin{itemize}%[topsep=.0in,leftmargin=0em,wide=0em,parsep=0em]
\item We provide a category theory perspective where we define category corresponding to the network structure and functors corresponding to the parameterization of the network.

\item From the category theory perspective, we further provide a concise definition for functionally equivalent features and formally define the feature complexity of a neural network. This differs largely from existing works dwelling on feature similarity.

\item Based on the definition of functionally equivalent feature, we theoretically prove that the linear mode connectivity , a widely-observed phenomenon in previous works, is a sufficient condition for the features learned by two different networks to be functionally equivalent.

\item Based on the metric of feature complexity, we further propose an algorithm called Iterative Feature Merging (IFM) to find functionally equivalent features in a trained neural network. Experimental results show its efficiency and potential as an effective pruning method. When taken as a pruning method, to the best of our knowledge, IFM is the first that neither requires fine-tuning nor requires manipulating the training of the network.

\item Empirical results of IFM yield interesting observations such as: 1) Larger networks tend to learn more functionally equivalent features. 2) Networks with the same structure but higher feature complexity achieve better performance. 3) Across the layers of a neural network, the feature complexity first increase then decrease. 4) In image classification tasks, a group of functionally equivalent features may correspond to a specific high-level semantic.
\end{itemize}

%\vspace{-10pt}
\section{Preliminaries}
%\vspace{-10pt}
\subsection{Notations}
%\vspace{-5pt}
Consider an $L$-layer neural network $f(\theta;\cdot)$ where $\theta$ is the parameter. We use $f^{l}(\theta^{l},\cdot)$ to denote the $l$-th layer. For the layer-wise feature map $Z^{l}(\theta, \mathbf{x})$, we have $Z^{l}(\theta, \mathbf{x}) = f^{l}(\theta^{l},Z^{l-1}(\theta, \mathbf{x}))$ and $Z^{1}(\theta, \mathbf{x}) = f^{1}(\theta^{1}, \mathbf{x})$. When $\mathbf{x}$ is not of concern, we abbreviate $Z^{l}(\theta)$ for $Z^{l}(\theta, \mathbf{x})$. 

To take a closer look, we proceed with an $L$-layer MLP for its ease of presentation, despite the fact that our definition of feature complexity is architecture agnostic and our proposed method could be applied to arbitrary architectures. For a MLP, the feature map $Z^{l}(\theta) \in \mathbb{R}^{d_l}$ contains $d_l$ features. We use $Z^{l}_{i}(\theta), i\in [1, d_l]$ to denote the $i$-the feature. The $\theta^{l}$ corresponds to a weight matrix $W^{l}\in \mathbb{R}^{d_l\times d_{l-1}}$ and a bias vector $\mathbf{b}^{l} \in \mathbb{R}^{d_l}$ and we have $Z^{l}(\theta) = \sigma(W^{l}Z^{l-1}(\theta) + \mathbf{b})$, where $\sigma$ corresponds to the activation function. For a permutation $\pi$ on $\theta$, we have $Z_{\pi}^{l}(\theta) = P_{l}Z^{l}(\theta)$ for each $l\in [1, L]$, where $P_{l}\in \mathbb{R}^{d_l\times d_l}$ is a permutation matrix. The weight is also permuted as $W_{\pi}^{l} = P_{l}W^{l}P_{l-1}^\top$ and $\mathbf{b}_{\pi}^{l} = P_{l}\mathbf{b}^{l}$, which make sure that $\forall \mathbf{x}, f(\theta, \mathbf{x}) = f(\pi(\theta), \mathbf{x})$.

%\vspace{-10pt}
\subsection{Category Theory}
%\vspace{-5pt}
\label{sec:CT}
Category theory is widely used in mathematics and many of the areas such as functional programming, which also attracts attention for neural networks like for foundation models~\cite{yuan2023power}. In this section, we introduce some basic and necessary concepts in category theory. For more comprehensive introduction please refer to \citet{mac2013categories,adamek1990abstract}.

\textbf{Category}~\citep{mac2013categories}: 
A category $C$ is a graph with a set of objects and a set of arrows (or ``morphisms"). For an arrow $f\in \mathbf{A}$ from object $a$ to object $b$, it is written as $f: a\rightarrow b$ where $a$ is the domain of $f$ written as $\texttt{dom} f = a$ and $b$ is the codomain of $f$ written as $\texttt{cod} f = b$. A category also have two additional operations:
\begin{itemize}
    \item \textit{Identity}: For each object $a$ there exists an arrow $id_{a}:a\rightarrow a$.
    \item \textit{Composition}: For each pair of arrows $<f,g>$ with $\texttt{dom} g = \texttt{cod} f$, there is an arrow $g\circ f:\texttt{dom} f \rightarrow \texttt{cod} g$ called their composite.
\end{itemize}

For the set of arrows from  $b$ to $c$ in category $C$, it is written as $Hom_{C}(a, b)$ called "hom-set".

\textbf{Functors}~\citep{mac2013categories}: 
A functor is the morphism between categories. For two category $C$ and $B$, a functor $T:C\rightarrow B$ consists two suitably related functions: the object function, which assigns each object $c$ of $C$ an object $Tc$ of $B$ and the arrow function which assigns each arrow $f:c\rightarrow c'$ of $C$ an arrow $Tf: Tc \rightarrow Tc'$ of $B$, in such way that:
\begin{equation}
    T(id_{c}) = id_{Tc}, \quad T(g\circ f) = Tg \circ Tf
\end{equation}
It means the mappings defined by the functor $T$ preserves the structure of the category $C$.

\textbf{Natural transformation}~\citep{mac2013categories}:
Given two functor $S,T: C\rightarrow B$, a natrual transformation $\tau:S\rightarrow T$ is a function assigns each object $c$ of $C$ an arrow $\tau_{c}: Sc\rightarrow Tc$ of $B$, in such way that every arrow $f:c\rightarrow c'$ of $C$ yields a commutative graph:
\begin{equation}
\label{eq:commute}
    \begin{adjustbox}
{center}
\begin{tikzpicture}
\draw (0,0) node at (2,2) {$Sc$} node at (4,2) {$Tc$} node at (2,0) {$Sc'$} node at (4,0) {$Tc'$};

\draw[thick, ->] (2.5,2) -- node[above] {$\tau_c$} (3.5,2);
\draw[thick, ->] (2.5,0) -- node[above] {$\tau_{c'}$} (3.5,0);
\draw[thick, ->] (2,1.5) -- node[left] {$Sf$} (2,0.5);
\draw[thick, ->] (4,1.5) -- node[right] {$Tf$} (4,0.5);
\end{tikzpicture}
\end{adjustbox}
\end{equation}
\vspace{-5pt}

Here commutative graph means that different paths between two objects yield, by composition, an equal arrow between the two objects, such that $Tf\left(\tau_{c}(Sc)\right) = \tau_{c'}\left(Sf(Sc)\right)$.

\section{Definition of Neural Network Feature Complexity}
%\vspace{-5pt}
In this section, we define the feature complexity of a neural network from a category theory perspective. We first introduce an abstract view of neural networks using the language of category theory (Sec.~\ref{sec:CT_NN}) and further define functionally equivalent feature from the category theory view (Sec.~\ref{sec:CT_LMC}). Finally, we provide the definition of feature complexity in Sec.~\ref{sec:Complexity}.
\vspace{-10pt}
\subsection{Category Theory View of Neural Networks}
\vspace{-5pt}
\label{sec:CT_NN}
For a neural network structure $f(\cdot, \cdot)$, it could be abstracted into a category $\mathcal{F}$ where objects are the shape of features and arrows are the type of transformation applied to the features, \emph{e.g.} linear transformation, convolution, attention \emph{etc.}. 
For an identity arrow, it simply apply no transformation on the feature while the composite of arrows is the composite of the corresponding transformations. 
Take a simple $L$-layer neural network structure for instance, it could be abstracted as the category depicted in Eq.~\ref{eq:simple_NN}, where each object corresponds to the shape of the input, feature maps and output while each arrow corresponds to the type of transformation applied to the feature (we omit identity circle and composites of arrows for simplicity).
\begin{equation}
    \mathbf{x} \mathop{\rightarrow}^{f^{1}} Z^{1} \mathop{\rightarrow}^{f^{2}} Z^{2} \mathop{\rightarrow}^{f^{3}} \cdots \mathop{\rightarrow}^{f^{L}} f(\cdot, x) .
    \label{eq:simple_NN}
\end{equation}

In the context of neural network structure, the training process aims to discover suitable parameters. In other words, it seeks to find a functor $T: \mathcal{F} \rightarrow \mathcal{P}$, where $\mathcal{P}$ represents a category describing the shapes of features, and its arrows encompass all possible transformations with specific parameters. For example, consider a network $f(\theta, \cdot)$ with parameter $\theta$, the corresponding functor $T_{\theta}$ maps each arrow $f^{l}(\cdot, \cdot)$ to $f^{l}(\theta^{l}, \cdot)$ with specific parameters $\theta^{l}$. To paint a clearer picture, as an analogy, we could think of the category $\mathcal{F}$ as a class in object-oriented programming language and think of the functor $T$ as creating an object instantiating the corresponding class.
%\vspace{-10pt}
\subsection{Definition of Functionally Equivalent Feature from the Category View}
%\vspace{-5pt}
\label{sec:CT_LMC}
Through the lens of category theory, as introduced in Sec.~\ref{sec:CT_NN},  we further define functionally equivalent feature using natural transformation (natural isomorphism). Consider a natural isomorphism $\tau$ 
between two functors $T_{\theta_a}, T_{\theta_b}: \mathcal{F}\rightarrow\mathcal{P}$ satisfying that for each object $z \in \mathcal{F}$, the transform $\tau_{z}:T_{\theta_a}z \rightarrow T_{\theta_b}z$ is an invertible transformation.
Natural transformations require naturality~\citep{mac2013categories}, ensuring that each arrow in category $\mathcal{F}$ results in a commutative graph in category $\mathcal{P}$, as described in Eq.~\ref{eq:commute}. Therefore, natural isomorphism between the functors regarding network parameters are non-trivial. 

\begin{definition}
\label{prop:feature_eq}
    \textbf{[Functionally Equivalent Feature]}
    For an $L$-layer neural network $f(\cdot, \cdot)$ and two different parameter $\theta_a$ and $\theta_b$, if there is a natural isomorphism between functor $T_{\theta_a}$ and $T_{\theta_b}$ then model $f(\theta_a, \cdot)$ and model $f(\theta_b, \cdot)$ have functionally equivalent features such that 
    \begin{equation}
        \forall \mathbf{x}\in \mathcal{D}, \forall l \in [1, L-1], \ \tau_{Z^{l+1}} Z^{l+1}(\theta_b, \mathbf{x}) = f^{l+1}\left(\theta_a^{l+1}, \tau_{Z^{l}}Z^{l}(\theta_b, \mathbf{x})\right).
    \end{equation}
    where $\tau_{Z^{l+1}}$ and $\tau_{Z^{l}}$ represent the invertible transformation defined by the natural isomorphism, and $Z^{l}(\theta_b, \mathbf{x})\in \mathbb{R}^{d_l}$ is the feature at the $l$-th layer.
\end{definition}

According to Definition~\ref{prop:feature_eq}, natural isomorphisms between networks indicate the features modeled by two networks are functionally equivalent \emph{i.e.} we could replace any feature maps of one model with the feature map from the other model through invertible linear transformation. Since it is a strong condition, simple transformations between different networks are unlikely to exist. 

Specifically, we notice linear mode connectivity (LMC)~\citep{frankle2020linear}, an empirical phenomenon that have drawn extensive attention. It says that there may exist a linear path between two different neural networks such that along the path, the loss is nearly constant. Various methods~\citep{entezari2022the,ainsworth2023git,DBLP:conf/icml/LiuLWXSY22} have been proposed to find networks satisfying LMC while recently a stronger notion of linear connectivity called Layerwise Linear Feature Connectivity (LLFC) was observed coexist with LMC~\citep{zhou2023going}.
We show that LMC actually indicates the functional equivalence between features.

\begin{proposition}
\label{prop:exist_simple_NT}
     \textbf{[LMC indicates functionally equivalent features]}
    (Proof in Appendix~\ref{app:proofs}).
    For two different parameter $\theta_a$ and $\theta_b$, if there is a permutation $\pi$ such that $\theta_a$ and $\pi(\theta_b)$ satisfy LMC, then there exists a natural isomorphism $\tau$ between functor $T_{\theta_a}$ and $T_{\theta_b}$.
\end{proposition}

Proposition~\ref{prop:exist_simple_NT} provides an interpretation of linear mode connectivity (LMC) from category theory perspective such that there always exists a natural isomorphism between two networks satisfying LMC, which indicates the two networks have functionally equivalent features. Note that \citet{entezari2022the,ainsworth2023git,DBLP:conf/icml/LiuLWXSY22} showed that different networks can be linearly connected after permutation, therefore we assume each $\tau_z$ to be permutation in the following, which means establishing a one to one correspondence between the features of the two networks.
%\vspace{-10pt}
\subsection{Definition of Feature Complexity based on Equivalent Feature}
%\vspace{-5pt}
\label{sec:Complexity}
Based on the definition of functionally equivalent feature, we could further define the feature complexity. 
Consider natural transformations mapping the functor $T_{\theta}$ to itself. The natural transformation that maps each arrow to itself must exist. In the case of other natural transformations, they map $Z^{l}(\theta)$ to $Z^{l}(\theta)P$ where $P\in \mathbb{R}^{d_l \times d_l}$ is a non-identity permutation matrix, which means mapping feature $Z^{l}_{i}(\theta)$ to $Z^{l}_{j}(\theta)$ with $i\neq j, i, j \in [1, d_l]$. This approach allows us to establish a partial order among features.
For two features $Z^{l}_{i}(\theta), Z^{l}_{j}(\theta), i,j \in [1, d_l]$ at $l$-th layer of network $f(\theta,\cdot)$, if there exists a simple transformation between $T_{\theta}$ to itself that maps $Z^{l}_{i}(\theta)$ to $Z^{l}_{j}(\theta)$, we say $Z^{l}_{i}(\theta) \leq Z^{l}_{j}(\theta)$.
It indicates that feature $Z^{l}_{j}(\theta)$ covers the feature represented in $Z^{l}_{i}(\theta)$ and could be used to replace $Z^{l}_{i}(\theta)$. It is obvious that if $Z^{l}_{i}(\theta)$ and $Z^{l}_{j}(\theta)$ is comparable then $Z^{l}_{i}(\theta) = Z^{l}_{j}(\theta)$.

\begin{theorem}
    \label{feature_comp}
    \textbf{[Feature Duality]} (Proof in Appendix~\ref{app:proofs}). 
    For an $L$-layer neural network $f(\theta,\cdot)$, there are multiple natural isomorphisms between $T_{\theta}$ to a itself if and only if 
    \begin{equation}
        \exists l\in [1, L], \exists i, j \in [1, d_l], \quad   s.t. \quad Z^{l}_{i}(\theta) = Z^{l}_{j}(\theta), i \neq j.
    \end{equation}
\end{theorem}

Finally, we formally define feature complexity based on the partial ordering between features.

\begin{definition}
    \label{def:feature_complexity}
    \textbf{[Feature Complexity]} Given an $L$-layer neural network $f(\theta,\cdot)$, the feature at $l$-th layer compose a poset $\{Z^{l}_{n}(\theta)| n\in [1, d_l]\}$. The maximum number of features that are not equivalent \emph{i.e.} the width of the corresponding poset is defined as the feature complexity at the $l$-th layer. 
\end{definition}

%In the partial ordered set $\{Z^{l}_{n}(\theta)| n\in [1, d_l]\}$ of features at $l$-th layer, each chain contains the functionally equivalent features that are comparable. Therefore, the width of the partial ordered set,  \emph{i.e.} the minimum number of chains at any partition or the cardinality of the maximum antichain , is the feature complexity. 

\begin{figure}[tb!]
%\vspace{-10pt}
\centering
    \subfigure[Weight matching distance between different features (normalized)]{
    \begin{minipage}[t]{0.45\textwidth}
        \centering   \includegraphics[width=1\linewidth]{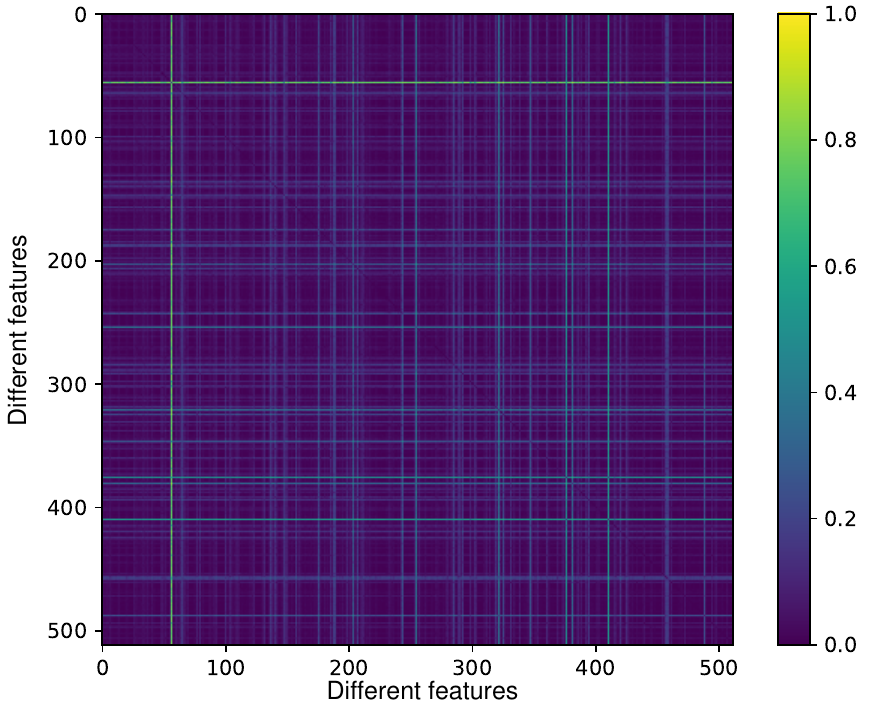}
    \end{minipage}
    \label{fig:norm_distance}
    }
    \subfigure[Testing accuracy on CIFAR10 of a model interpolating between a VGG16 and the same model after permutation]{
    \begin{minipage}[t]{0.51\textwidth}
        \centering \includegraphics[width=1\linewidth]{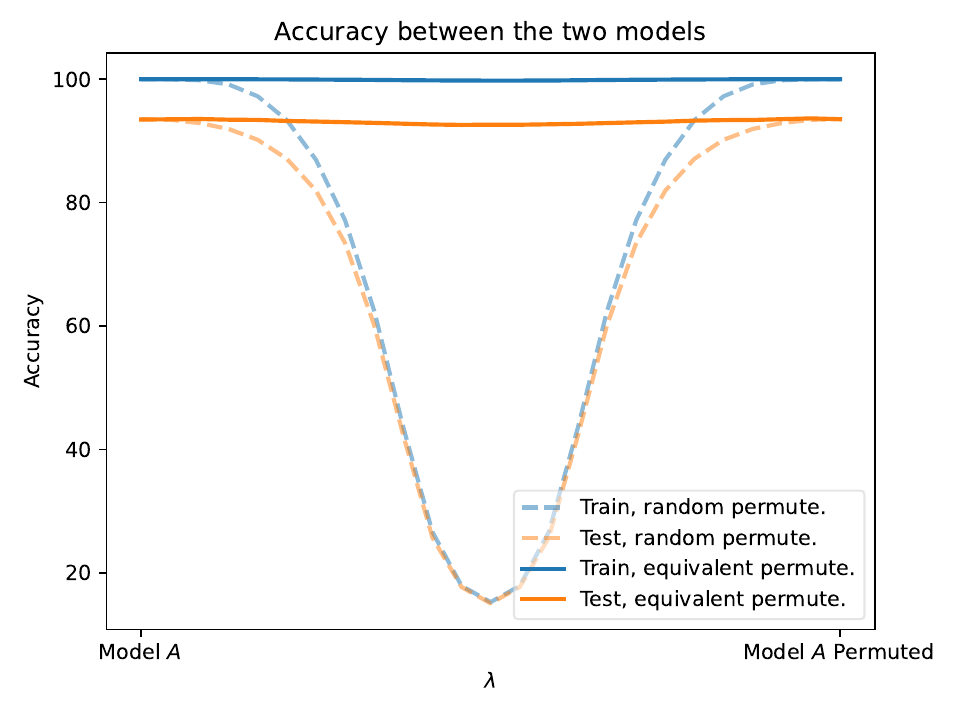}
    \end{minipage}
    \label{fig:interp_permute}
    }
    \label{fig:existence}
    \vspace{-10pt}
    \caption{The empirical evidences that functionally equivalent features exist. (a): The heatmap of the distances between different features from the last convolution layer of VGG16 on CIFAR10. (b):The test accuracy on CIFAR10 of a model interpolating between a VGG16 and the same model after permutation on each layer. The solid lines are the result of permutation on functionally equivalent features while the dashed lines are the result of random permutation on the same number of features.}
    %\vspace{-5pt}
\end{figure}
%\vspace{-10pt}
\section{Measuring Neural Network Feature Complexity}
%\vspace{-5pt}
\label{sec:measure}
In this section, we propose an algorithm to empirically measure the complexity of the features by finding equivalent features with weight matching and merging the equivalent features. Firstly, we devise and introduce two components to find and merge equivalent features.

\textbf{Feature weight matching: }
Drawing inspiration from the weight matching method used to identify permutations in linear mode connectivity literature, we propose matching the weights corresponding to each feature. 
%Suppose the feature of network $f(\theta, \cdot)$ at $l$-th layer is $Z^{l}\in \mathbb{R}^{d_l}$ where $Z^{l}_m \in \mathbb{R}, m\in [1, d_l]$ repersents the $m$-th feature and the weight at $l$-th layer is $W^{l} \in \mathbb{R}^{d_l \times d_{l-1}}$ where $d_l$ and $d_{l-1}$ is the number of the features in $l$-th layer and $l-1$-th layer. 
Denote $W^{l}_{[i,:]}$ as the $i$-th row of $W^{l}$ and $W^{l}_{[:,j]}$ as the $j$-th column of $W^{l}$. For two features $Z^{l}_m, Z^{l}_n, m,n \in [1, d_l]$, the weight distance between the two features is defined as:
\begin{equation}
    \label{eq:distance}
    D^{l}_{mn} = \|W^{l}_{[m,:]} - W^{l}_{[n,:]} \|^2 + \|W^{l+1}_{[:,m]} - W^{l+1}_{[:,n]} \|^2
\end{equation}
When matching the weights, we consider the layers such as linear fully connected layers or convolution layers at the $l$-th layer, while ignoring the activation and normalization layers in the network.

\textbf{Feature merging:} Suppose we find two functionally equivalent features $Z^l_m, Z^l_n$ at the $l$-th layer, where $m,n\in [1,d_l]$. To obtain a merged feature, we merge the weight at the $l$-th layer. For the merged weight at $l$-th layer $W^{\prime l}\in \mathbb{R}^{(d_l-1)\times d_{l-1}}$, the row of weight matrix $W^{\prime l}$ corresponding to the merged feature is:
\begin{equation}
    \label{eq:merge_wl}
    W^{\prime l}_{merged} = W^{l}_{[m,:]} + W^{l}_{[n,:]}
\end{equation}

The other $d_l-2$ rows of $W^{\prime l}$ correspond to the $d_l-2$ rows in $W^{\prime l}$ such as $W^{l}_{[i,:]}, i\neq m, n$.

Therefore the merged feature at the $l$-th layer becomes $Z^{\prime l}\in \mathbb{R}^{d_l-1}$ where the merged feature $Z^{\prime l}_{d_l-1} = Z^{l}_m + Z^{l}_n$ and the other $d_l-2$ features correspond to $Z^l_{i}, i\neq m,n$.

To process $Z^{\prime l}$, we adjust the weight at $l+1$-th layer. For $W^{\prime (l+1)}\in \mathbb{R}^{d_{l+1}\times d_l-1}$, the column of $W^{\prime (l+1)}$ corresponding to the merged feature is:
\begin{equation}
\label{eq:merge_wl+1}
    W^{\prime (l+1)}_{merged} = \mathop{mean}\left(W^{l+1}_{[:, m]}, W^{l+1}_{[:,n]}\right)
\end{equation}

The other $d_l-2$ columns of the weight $W^{\prime (l+1)}$ corresponds to the $d_l-2$ columns $W^l_{[:,j]}, j\neq m,n$.

Using this method, we could merge features to obtain an approximately functionally equivalent neural network. Note that when merging several features the process is similar to Eq.~\ref{eq:merge_wl} and Eq.~\ref{eq:merge_wl+1}.

\textbf{Iterative Feature Merging with Weight Matching:} Composing the two components proposed above, we get the algorithm to measure the feature complexity. We iteratively merge the features in each layer $l$ with the smallest weight distance defined in Eq.~\ref{eq:distance} until 
\begin{equation}
    \min_{m,n\in [1, d_l], m\neq n} D^l_{mn} > \beta \max_{m,n\in [1, d_l], m\neq n} D^l_{mn}
\end{equation}
In this context, $\beta$ represents a hyper-parameter, and $D^{l}_{mn}$ corresponds to the weight distance as defined in Eq.~\ref{eq:distance}.

To get the mean value in Eq.~\ref{eq:merge_wl+1}, we also keep track of the number of features merged into one feature. Consider merging two feature $Z^{l}_m$ and $Z^{l}_n$ where $Z^{l}_m$ is the merged feature of $N_m$ features and $Z^{l}_n$ is the merged feature of $N_n$ features, then we have:
\begin{equation}
    \label{eq:merge_l+1_v2}
    W^{\prime (l+1)}_{merged} = (N_{m_{min}}W^{l+1}_{[:, m_{min}]} + N_{n_{min}}W^{l+1}_{[:,n_{min}]})/(N_{m_{min}} + N_{n_{min}})
\end{equation}

The detailed algorithm for Iterative Feature Merging (IFM) is presented in Algorithm~\ref{alg:IFM} in Appendix~\ref{app:algorithm}.

%\vspace{-10pt}
\section{Experiments}
%\vspace{-5pt}
We conduct experiments to verify the effectiveness and efficiency of our proposed iterative feature merging (IFM) algorithm. In Sec.~\ref{sec:exp_exist}, we empirically verify the existence of functionally equivalent features. In Sec.~\ref{sec:exp_pruning}, we show that the iterative feature merging could greatly reduce the number of parameters while maintaining the performance without fine-tuning. In Sec.~\ref{sec:exp_empirical}, we provide more empirical results regarding the feature complexity. The experiments are based on VGG and ResNet models on CIFAR10 and ImageNet, respectively. For more details, refer to Appendix~\ref{app:exp_detail}.
%\vspace{-10pt}
\subsection{Verifying the Existence of Equivalent Features}
%\vspace{-5pt}
\label{sec:exp_exist}
In this section, we empirically show that the functionally equivalent feature as defined in Sec.~\ref{sec:CT_LMC} actually exist in vanilla networks. We conduct experiments on VGG16 trained on CIFAR10 with SGD optimizer for $150$ epochs. For details, refer to Appendix~\ref{app:exp_detail}.

As shown in Fig.~\ref{fig:norm_distance}, we visualize the normalized distances between $512$ different features from the last convolution layer of the VGG16. The minimum distance between two features is $2e-15$. For most features, their distance is relatively low. Similar results are observed for other layers. For more results please refer to Appendix~\ref{app:add_results}.

For a model $f(\theta, \cdot)$, we further consider the permutation $\pi$ that swaps the functionally equivalent features found by the feature weight matching introduced in Sec.~\ref{sec:measure}.
Similar to the metric used to identify LMC, we further interpolate between the parameter $\theta$ and the one after permutation $\pi(\theta)$ and test the accuracy of model $f\left((1 - \lambda) \theta + \lambda \pi(\theta), \cdot\right)$. The results are shown in Fig.~\ref{fig:interp_permute} where solid lines correspond to the permutation swapping equivalent features and dashed lines correspond to random permutation swapping the same number of features for comparison. The results show that interpolating between functionally equivalent features hardly affects the performance.
\begin{figure}[tb!]
%\vspace{-10pt}
\centering
    \subfigure[Testing accuracy to the percentage of remaining parameters on VGG networks]{
    \begin{minipage}[t]{0.48\textwidth}
        \centering   \includegraphics[width=1\linewidth]{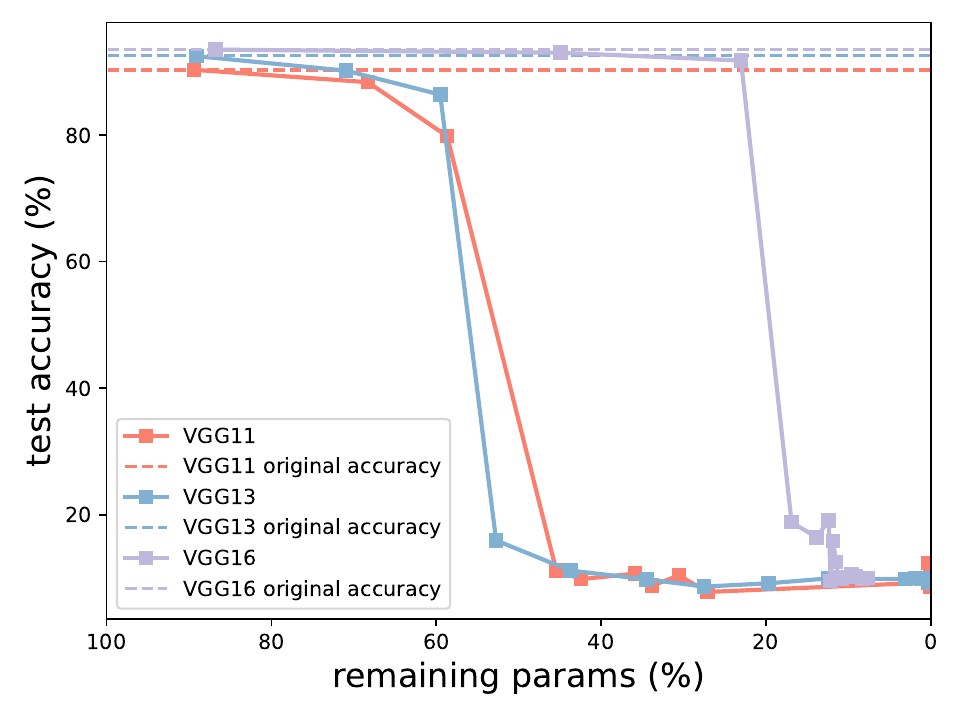}
    \end{minipage}
    \label{fig:vgg_CIFAR10}
    }
    \subfigure[Testing accuracy to the number of remaining parameters on VGG and ResNet]{
    \begin{minipage}[t]{0.48\textwidth}
        \centering \includegraphics[width=1\linewidth]{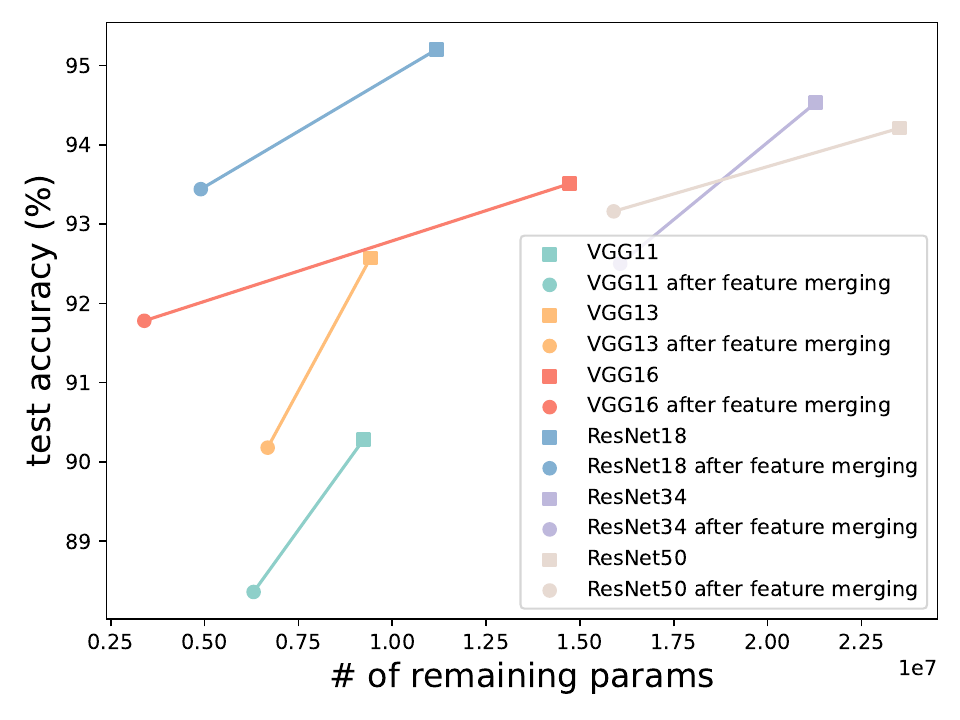}
    \end{minipage}
    \label{fig:resnet_vgg_CIFAR10}
    }
    \label{fig:CIFAR10_ifm}
    \vspace{-10pt}
    \caption{Results for iterative feature merging on CIFAR10. (a): Testing accuracy to the percentage of remaining parameters. Each dot corresponds to a different hyper-parameter $\beta$ in iterative feature merging. (b): Testing accuracy to the number of parameters. We conduct a grid search on $\beta$ and choose the largest $\beta$ with testing accuracy larger than $95\%$ of the testing accuracy before merging.}
    %\vspace{-10pt}
\end{figure}

%\vspace{-10pt}
\subsection{Iterative Feature Merging}
%\vspace{-5pt}
\label{sec:exp_pruning}
In this section, we present the results of iterative feature merging with VGG~\citep{SimonyanZ14aiclr} and ResNet~\citep{He16cvpr} on CIFAR10~\citep{krizhevsky2009learning} and ImageNet~\citep{deng2009imagenet}. We train models on CIFAR10 and use pretrained checkpoints on ImageNet provided by torchvision~\citep{TorchVision}. For details, refer to Appendix~\ref{app:exp_detail}.

\subsubsection{Iterative Feature Merging on CIFAR10}

With a varying $\beta$ in Algorithm~\ref{alg:IFM}, the testing accuracy would change accordingly, as different number of features are merged.
The results for VGG networks on CIFAR10 are shown in Fig.~\ref{fig:vgg_CIFAR10}, with the x-axis representing the percentage of remaining parameters and the y-axis representing testing accuracy. \textbf{It is clear that the larger the network is, the more equivalent features could be merged until the reduction eventually significantly affects the performance.} Specifically, for VGG16, we could reduce the number of parameters to $23.03\%$ while keeping the testing accuracy at $91.78\%$ which is relatively close to the original accuracy at $93.51\%$.

We further demonstrate the relationship between testing accuracy and the number of network parameters in Fig.~\ref{fig:resnet_vgg_CIFAR10}. Here we apply grid search on $\beta$ and choose the largest $\beta$ with the testing accuracy larger than $95\%$ of the testing accuracy before feature merging. \textbf{With iterative feature merging, larger models (\emph{e.g.} VGG16 and ResNet50) could reduce more parameter with a smaller decrease on testing accuracy.} Note that we also find ResNet18 outperforming ResNet34 and ResNet50 on CIFAR10. For the reason of this phenomenon, we believe that the capacity of ResNet18 is already enough to deal with the simple classification task on CIFAR10, \emph{e.g.} we could reduce up to $60\%$ of the parameters of ResNet-18 without significantly affecting the testing accuracy. 
%(from $95.20\%$ to $93.44\%$).

\begin{figure}[tb!]
%\vspace{-10pt}
\centering
    \subfigure[Testing accuracy to the percentage of remaining parameters on VGG networks]{
    \begin{minipage}[t]{0.48\textwidth}
        \centering   \includegraphics[width=1\linewidth]{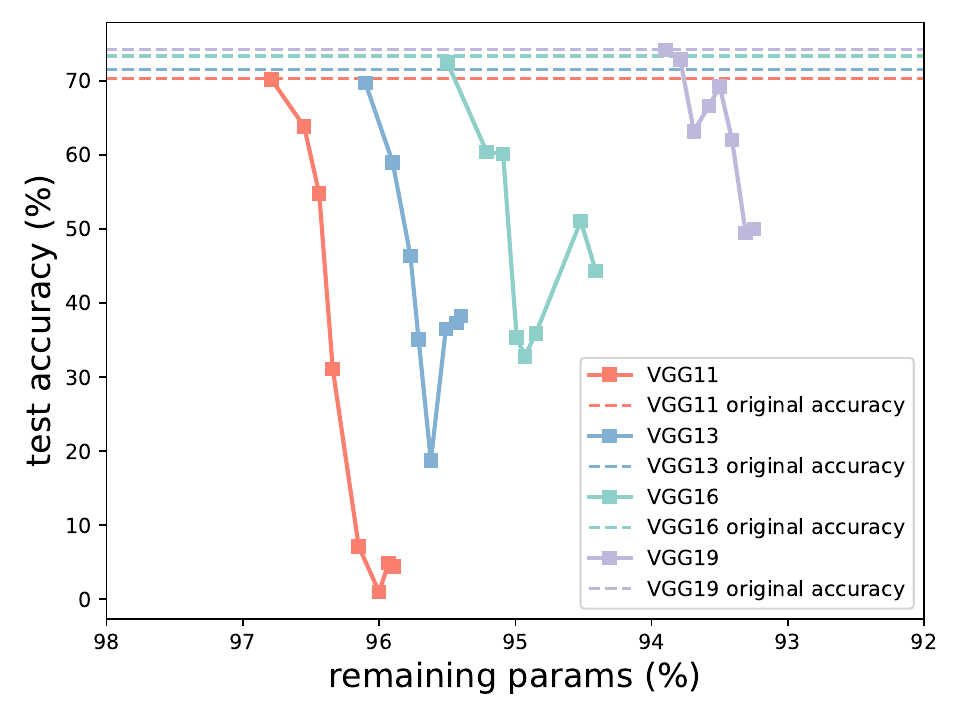}
    \end{minipage}
    \label{fig:vgg_imagenet}
    }
    \subfigure[Testing accuracy to the percentage of remaining parameters on ResNet50 with two different parameters]{
    \begin{minipage}[t]{0.48\textwidth}
        \centering \includegraphics[width=1\linewidth]{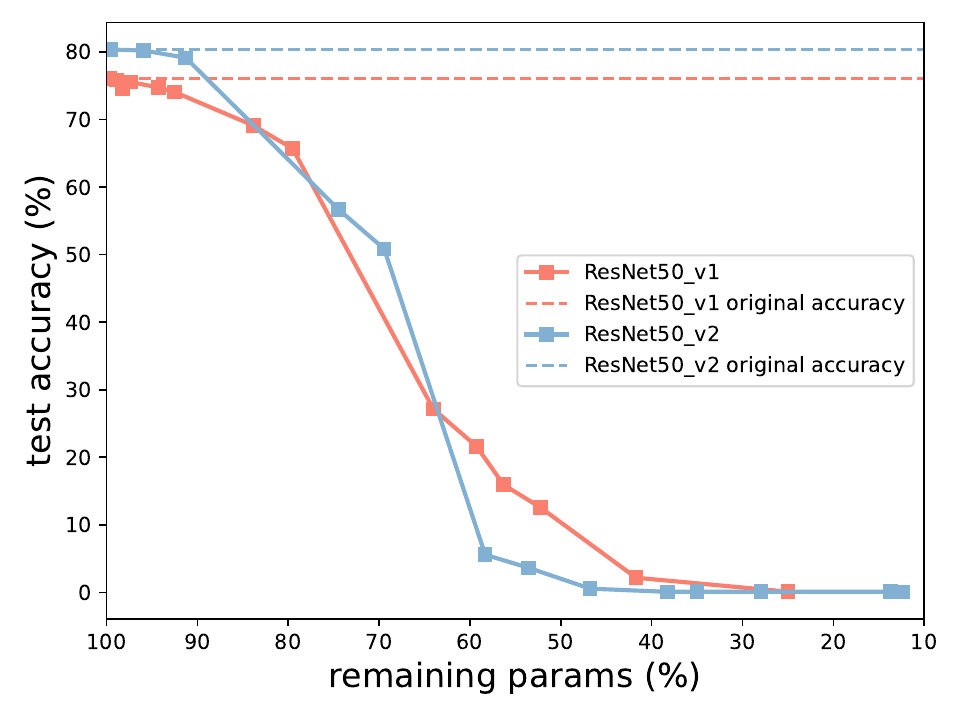}
    \end{minipage}
    \label{fig:resnet_imagenet}
    }
    %\vspace{-10pt}
    \caption{Results of our iterative feature merging on ImageNet. (a): Testing accuracy (top-1) in relation to the percentage of remaining parameters of VGGs after feature merging with different $\beta$. (b): Testing accuracy (top-1) to the percentage of remaining parameters of ResNet50 after feature merging. ``ResNet50\_v1" and ``ResNet50\_v2" corresponds to the two different checkpoints in torchvision, respectively.}
    %\vspace{-10pt}
\label{fig:imagenet_result}
\end{figure}

\subsubsection{Iterative Feature Merging on ImageNet}

ImageNet is a much larger and more complex dataset compared to CIFAR10. \textbf{We find that neural networks learn more complex features to solve the more difficult task, \emph{i.e.} image classification on ImageNet.} As shown in Fig.~\ref{fig:imagenet_result}, iterative feature merging only reduce approximately $5\%$ parameters for VGG and approximately $10\%$ parameters for ResNet50. 

We observe the same trend as on CIFAR10 that the larger the network is, the more equivalent features could be merged, as shown in Fig.~\ref{fig:vgg_imagenet}. For different parameters with the same network structure, we conduct experiments on two different parameters of ResNet50 in torchvision, where "ResNet50\_v2" achieves higher top-1 accuracy than "ResNet50\_v1" ($80.86\%$ to $76.13\%$) due to different training hyper-parameters. As shown in Fig.~\ref{fig:resnet_imagenet}, the testing accuracy of ResNet50\_v2 decreases faster than the testing accuracy of ResNet50\_v1 as more features are merged. \textbf{It indicates that ResNet50\_v2 learns more diverse features, which may explain the superior performance of ResNet50\_v2.}

\begin{wrapfigure}{r}{0.43\textwidth}
    \centering
    %\vspace{-8pt}
    \includegraphics[width=0.95\linewidth]{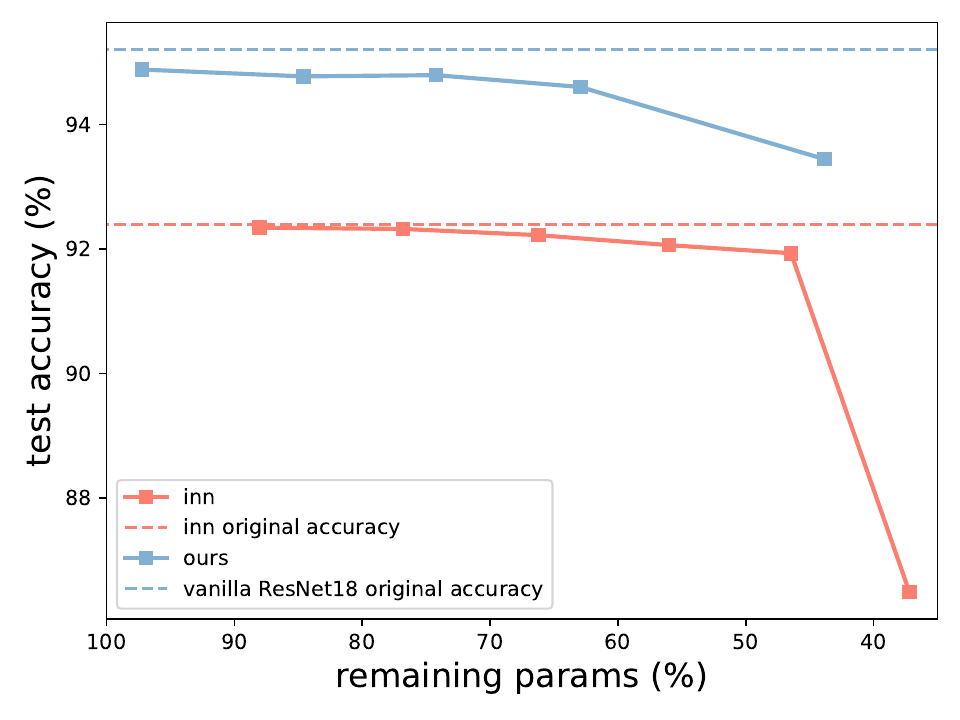}
    %\vspace{-8pt}
    \caption{Pruning results of ResNet18 on CIFAR10. The INN method  are evaluated with the released official code.}
    %\vspace{-10pt}
    \label{fig:compare}
\end{wrapfigure}

\subsubsection{The Potential of IFM for Pruning}
IFM merges functionally equivalent features, which is similar to the channel pruning~\citep{Wen16nips,Ye18iclr} that remove channels to reduce computational cost. Note that conventional pruning methods require fine-tuning the pruned model while our proposed method does not. In Fig.~\ref{fig:compare}, we compare our IFM with a recently SOTA pruning method ``INN"~\citep{Solodskikh23cvpr} which also does not require the fine-tuning. It shows that our IFM keeps a higher accrucay over different pruning ratio for ResNEt18 on CIFAR10. Note that INN also requires changing the training procedure of neural networks. In contrast, our IFM, to the best of our knowledge, is the first non-intrusive prune method does not require access to the training data. 
It can be directly applied to vanilla pre-trained models with high efficiency and simplicity. Please refer to Appendix~\ref{app:add_results} for the time complexity analysis. However, it is also observed that IFM is less effective on large-scale datasets such as ImageNet which we analyze is due to the complexity of the dataset which calls for higher feature complexity for neural networks to solve the task. 
%However, it is worth noting that the pruning effect of IFM may be limited when applied to complex tasks such as image classification on ImageNet since we observed that the feature complexity is higher on neural networks trained on ImageNet comparing to neural networks trained on CIFAR10.
%Our IFM also have limitations such that the parameter reduction may be limited on complex datasets \emph{e.g.} ImageNet.
%\vspace{-10pt}
\subsection{Empirical Results on Feature Complexity}
\begin{figure}
\vspace{-12pt}
    \centering
    \subfigure[Feature complexity at each layer of VGGs]{
    \begin{minipage}[t]{0.45\textwidth}
        \centering   \includegraphics[width=1\linewidth]{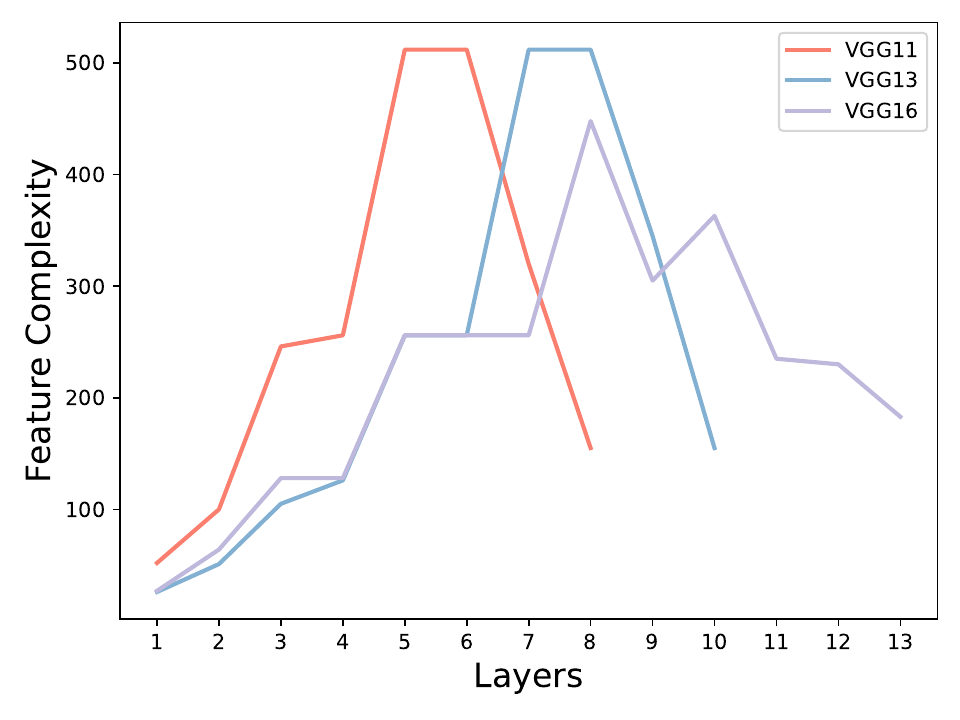}
    \end{minipage}
    \label{fig:vgg_compelxity}
    }
    \hspace{2pt}
    \subfigure[Guided backpropagation results]{
    \begin{minipage}[t]{0.48\textwidth}
        \centering \includegraphics[width=1\linewidth]{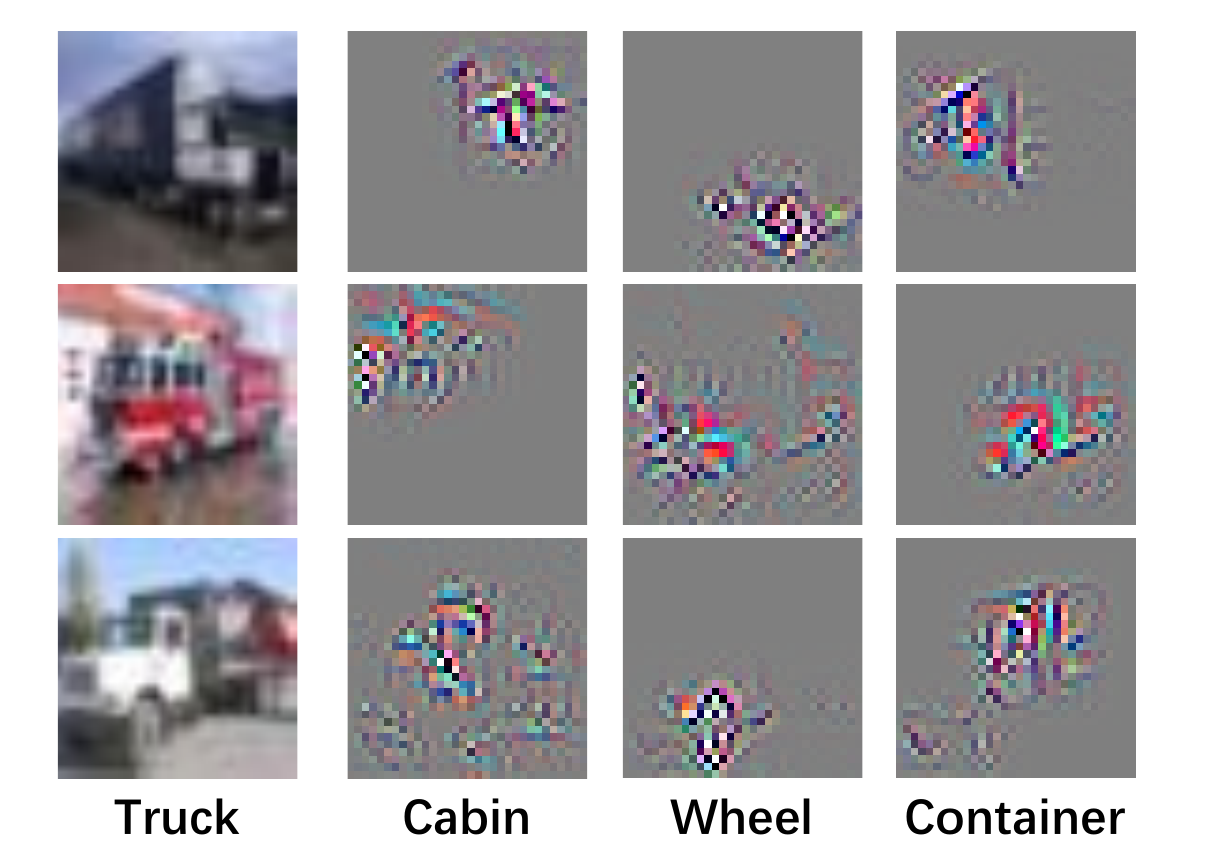}
    \end{minipage}
    \label{fig:backprop}
    }
    
  %  \vspace{-10pt}
    \caption{Empirical results regarding feature complexity. (a): The plot of feature complexity at each layer of VGG networks that is trained on CIFAR10. (b): The gradient results on the raw images by the guided backpropagation of three different sets of equivalent features in the VGG16 on three different images of truck from CIFAR10.}
  %  \vspace{-15pt}
\label{fig:empirical}
\end{figure}

%\vspace{-5pt}
\label{sec:exp_empirical}
In this section, we present some empirical results regarding the feature complexity. In Fig.~\ref{fig:vgg_compelxity}, we present the feature complexity at each layer of the VGG networks trained on CIFAR10. \textbf{We find that the feature complexity is increasing at the first several layers and decreasing at the last several layers while the maximum feature complexity is reached in the middle.} Several possible reasons could be used to explain this phenomenon, \emph{e.g.} the neural network starts to forget features or many simple features compose complex features at the last several layers. We leave the explanation of this phenomenon for future research.

In Fig.~\ref{fig:backprop}, we visualize the guided backpropagation of various features in VGG16. Guided backpropagation~\citep{Springenberg14iclr} is a method to visualize the part of the image that activate neurons in the neural network by applying ReLU on the gradient through backpropagation. Here we visualize the guided backpropagation of a group of functionally equivalent features by setting the gradient of other features to be zero. As shown in Fig.~\ref{fig:backprop}, a group of functionally equivalent features capture the similar features on different images. The three different groups of functionally equivalent features we demonstrate capture the cabin, wheel and container of trucks respectively. \textbf{It suggests the possibility to align semantics to a group of equivalent features learned by the neural network for classification.} Please refer to Appendix~\ref{app:add_results} for more results.

%\subsection{Time Complexity of Iterative Feature Merging}
%\label{sec:exp_time_complexity}
%、vspace{-10pt}
\section{Related Works}
%\vspace{-5pt}
\textbf{Feature Similarity.} To understand and harness the behaviour of neural networks during training andd testing, various efforts have been made especially for studying the features (representations) learned by networks including the transferability of features~\citep{Yosinski14nips}, the training dynamics~\citep{Morcos18nips}, the effect of network width and depth on the learned feature~\citep{Nguyen21iclr}. Among these works, measuring the feature similarity (representational similarity) has become an important issue. Various metrics have been proposed to measure the feature similarity including the measures based on canonical correlation analysis~\citep{RaghuGYS17nips,Morcos18nips,GodfreyBEK22NIPS}; those based on alignment~\citep{li2015convergent, Williams21nips}; those based on representational similarity matrix~\citep{Shahbazi21, Tang20arxiv} and those based on topology~\citep{Barannikov22ICML} \emph{etc.}  Based on various feature similarity measures, previous works have shown that there are similar features within a network with increased width and depth~\citep{Nguyen21iclr} and different networks with similar performance learn similar features~\citep{li2015convergent, Wang18nips}. However, similar features could be vague considering various similarity measures. In this paper, instead of focusing on similar features, we formally define functionally equivalent features and provide actionable insights such that neural networks learn redundant features that could be merged without affecting performance.

\textbf{Linear Mode Connectivity.} \citet{freeman2017topology, draxler2018essentially, garipov2018loss} observed Mode Connectivity, i.e., different well-trained models can be connected through a non-linear path of nearly constant loss. 
\citet{frankle2020linear} first proposed the notion of Linear Mode Connectivity (LMC), where models are connected through linear path of constant loss. \citet{frankle2020linear} observed LMC for networks that are jointly trained for a short amount of time before going through independent training (referred as the {spawning method}). 
Later, \citet{entezari2022the,ainsworth2023git,DBLP:conf/icml/LiuLWXSY22} showed that even independently trained networks can be linearly connected when permutation invariance is taken into account. Permutation invariance of NNs indicates one can permute neurons of each layer of NNs without changing their functionality. 
In particular, \citet{ainsworth2023git} utilized the permutation invariance to align the neurons of two neural networks and formulate the neuron alignment problem as a bipartite graph matching problem.  \citet{ainsworth2023git} proposed two matching methods: {activation matching} and {weight matching}, referred as the {permutation methods}.
More recently, \citet{zhou2023going} discovered a stronger notion of linear connectivity, called Layerwise Linear Feature Connectivity (LLFC), which indicates that the feature maps of every layer in different trained networks are also linearly connected, and demonstrated the co-occurrence of LLFC and LMC. In this paper, we show that LMC or LLFC between two different networks indicates the two networks learn functionally equivalent features.

\textbf{Category Theory in Machine Learning} Category theory have been used in machine learning~\citep{shiebler2021category} for various topics including backpropagation~\citep{fong2019backprop}, categorical probabilities~\citep{fritz2020synthetic},  conditional independence~\citep{mahadevan2022categoroids}, 
supervised learning~\citep{harris2019characterizing},
reinforcement learning~\citep{mahadevan2022unifying} and so on. Among them, \citet{bradley2022enriched} provides an enriched category theory of natural language while \citet{yuan2023power} provides an analysis upon the power of perfect foundation models from category theory perspective. These previous works define different categories and functors from different perspective. In this paper, we provide a new perspective by defining the category regarding the network structure and the functor corresponding to the parameterization of the network. 

\textbf{Model Compression and Network Pruning} The increasing size of neural networks have motivated the standing research on network compression over the decade. To reduce the computation cost, various methods have been proposed: knowledge distillation~\citep{Hinton15}, quantization~\citep{gong2014compressing,wu2016quantized}, low-rank factorization~\citep{tai2015convolutional}, \emph{etc.} 
Specifically, network pruning~\citep{hanson1988comparing} tend to reduce redundant parameters that are not sensitive to the performance. The pruning methods include unstructured pruning~\citep{lecun1989optimal,hassibi1993optimal,srinivas2015data} and structured pruning~\citep{zhou2016less,Wen16nips,Ye18iclr}. The unstructured pruning remove connections between neurons that the dimensionality of the feature map does not change. The structured pruning, on the other way, removes groups of weights (filters, channels, \emph{etc.}) and reduces the dimension of the feature map. Our proposed IFM could also be considered as a structured pruning method. Note that most pruning methods require access to data \emph{e.g.} fine-tuning, retraining \emph{etc.} Efforts have been made to develop data agnostic prune methods. For pruning at initialization (PaI)~\citep{wang2021recent}, \citet{tanaka2020pruning} propose a data agnostic method to prune the network before training. For pruning after training (PaT), \citet{Solodskikh23cvpr} propose a new training algorithm that does not require fine-tuning to prune the model after training. However, they either require the access to training (fine-tuning) procedure after pruning or before pruning. To the best of our knowledge, the IFM is the first data-agnostic prune method not requiring any training or fine-tuning.

\begin{table}[tb!]
    \centering
        \caption{Comparison between popular pruning methods about whether they require access to data before and after pruning. Note that our method requires no data both before and after pruning.}
     \resizebox{\textwidth}{24mm}{
    \begin{tabular}{c|c|c}
    \toprule
       Method  & access to data before pruning & training (fine-tuning) after pruning\\
       \hline
       \makecell[l]{Structured Sparsity Learning (SSL)\\
       \citep{Wen16nips}} &  $\checkmark$ & $\times$ \\
       \hline
       \makecell[l]{Iterative Magnitude Pruning(IMP) \\ \citep{FrankleC19iclr,frankle2019stabilizing}}   &  $\checkmark$ & $\checkmark$ \\
       \hline
       \makecell[l] {SNIP\\
       \citep{Lee19ICLR}} & $\checkmark$ & $\checkmark$\\
       \hline
       \makecell[l]{Iterative Synaptic Flow Pruning (SynFlow) \\
       \citep{tanaka2020pruning}} & $\times$ & $\checkmark$\\
       \hline
       \makecell[l]{Integral Neural Network (INN)\\
       \citep{Solodskikh23cvpr}} & $\checkmark$ & $\times$\\
       \hline
       \makecell[l]{Iterative Feature Merging (IFM)\\ (Ours)} & $\times$ & $\times$\\
    \bottomrule
    \end{tabular}}
    \label{tab:prune_compare}
\end{table}

\section{Conclusion and Outlook}
%\vspace{-5pt}
In this paper, we have tried to formally define functionally equivalent features and further introduce the concept of feature complexity, an inherent measure of the complexity of features learned by the neural network. To measure the feature complexity, we propose an efficient algorithm called Iterative Feature Merging. Experiments on CIFAR10 and ImageNet have shown its effectiveness and  insights regarding the feature complexity can be derived. The proposed IFM also shows potential in pruning pre-trained models without fine-tuning which is often needed in existing pruning methods~\citep{Wen16nips,Ye18iclr}.

There are much room for further improvement and extension in future work. For example, since in current work we do not proactively modify the structure and training procedure of the neural network, when facing difficult tasks (e.g. ImageNet classification), the number of parameters that could be reduced by IFM is limited. For better pruning performance, designing network structure and training algorithms to learn a more compact representation could be a promising direction, which we leave for future researches. Though our proposed definition and method are network structure and data agnostic, another major limitation is that we only conduct experiments for image classification task. More experiments could be done regarding different network structures and different tasks. The functionally equivalent feature defined in this paper could also be used as a tool to analyze the neural networks under out-of-distribution scenarios or adversarial attacks. In addition, we may also apply our theory and methods on the binary or ternary weight networks~\citep{LiuTWN22}, or the addernet~\citep{addernetCVPR20} that reduces the multiplication operations, to see if there is still space to reduce the model. It is also attractive to apply our work to graph neural networks (GNNs)~\citep{gori2005new,scarselli2008graph,bruna2013spectral,kipf2016semi,KipfW17iclr,WuYZHWY23ICLR} especially considering their relative smaller size of training data.

We hope that our proposed feature complexity will inspire future research for a better understanding the behavior and enhance the performance of neural networks.

\bibliography{iclr2024_conference}
\bibliographystyle{iclr2024_conference}

\newpage
\appendix
\section{Algorithm for Iterative Feature Merging}
\label{app:algorithm}
In this section, we present the detailed algorithm for iterative feature learning (IFM) in Algorithm~\ref{alg:IFM}.

%\vspace{-5pt}
\begin{algorithm}[htb!]
\caption{Iterative Feature Merging with Weight Matching}
\label{alg:IFM}
\begin{algorithmic}
\STATE {\bfseries Input:} Parameter $\theta$ for an $L$-layer neural network $f(\cdot, \cdot)$, where the weight at the $l$-th layer is $W^{l}\in \mathbb{R}^{d_l\times d_{l-1}}$; hyperparameter $\beta$
\FOR{$l \in [1, L]$}
\WHILE{True}
    \FOR{$m \in [1, d_l]$}
        \STATE $N_{m} \leftarrow 1$
        \FOR{$n \in [1, d_l]$}
        \STATE $D^l_{mn} \leftarrow \|W^{l}_{[m,:]} - W^{l}_{[n,:]} \|^2 + \|W^{l+1}_{[:,m]} - W^{l+1}_{[:,n]} \|^2$
        \ENDFOR
    \ENDFOR
    \IF{$\min_{m\neq n} D^{l}_{mn} > \beta \max_{m\neq n} D^{l}_{mn}$}
    \STATE break
    \ELSE
    \STATE $m_{min}, n_{min} \leftarrow \arg \mathop{\min}_{m,n, m\neq n} D^{l}_{mn}$ 
    \STATE $W^{\prime l}_{merged} \leftarrow W^{l}_{[m_{min}, :]} + W^{l}_{[n_{min}, :]}$
    \STATE $W^{\prime (l+1)}_{merged} \leftarrow (N_{m_{min}}W^{l+1}_{[:, m_{min}]} + N_{n_{min}}W^{l+1}_{[:,n_{min}]})/(N_{m_{min}} + N_{n_{min}})$
    \STATE $N_{merged} \leftarrow N_{m_{min}} + N_{n_{min}}$
    \ENDIF
\ENDWHILE
\ENDFOR
\end{algorithmic}
%\vspace{-2pt}
\end{algorithm}

\section{Proofs}
\label{app:proofs}
\subsection{Proof for Proposition~\ref{prop:exist_simple_NT}}
Let us first formally define linear mode connectivity.

\begin{definition}
\label{def:lmc}
\textbf{[Linear Mode Connectivity]}
Given a dataset $\mathcal{D}$ and two neural networks of the same structure and different parameters $\theta_a$ and $\theta_b$ with similar loss $\mathcal{L}_{\mathcal{D}}(\theta_a) \approx \mathcal{L}_{\mathcal{D}}(\theta_b)$, the two networks are linearly connected if
\begin{equation}
    \mathcal{L}_{\mathcal{D}}(\theta_a) \approx \mathcal{L}_{\mathcal{D}}(\theta_b) \approx 
    \mathcal{L}_{\mathcal{D}}\left(\alpha\theta_a + (1-\alpha)\theta_b\right); \quad \forall \alpha \in [0, 1].
\end{equation}
\end{definition}

Beyond LMC, the linear layer-wise feature connectivity (LLFC) was observed coexist with LMC.

\begin{definition}
    \label{def:llfc}
    \textbf{[Layerwise Linear Feature Connectivity]}
    Given a dataset $\mathcal{D}$ and two $L$-layer neural networks of the same structure and different parameters $\theta_a$ and $\theta_b$. They satisfy LLFC if
    \begin{equation}
    \label{eq:B_1_1}
        \forall \mathbf{x} \in \mathcal{D}, \forall l \in [1, L], \forall \alpha \in [0,1], Z^{l}\left(\alpha\theta_a + (1-\alpha)\theta_b, \mathbf{x}\right) = \alpha Z^{l}(\theta_a, \mathbf{x}) + (1-\alpha) Z^{l}(\theta_b, \mathbf{x}) .
    \end{equation}
\end{definition}

\begin{proof}
    \textbf{[Proof for Proposition~\ref{prop:exist_simple_NT}]}
    Consider two parameter $\theta_a$ and $\theta_b$ satisfying LMC, according to Eq.~\ref{eq:B_1_1} we have
    \begin{equation}
    \begin{split}
    \label{eq:B_1_2}
        f^{l}\left(\alpha\theta_a + (1-\alpha)\theta_b, \alpha Z^{l-1}(\theta_a, \mathbf{x}) + (1-\alpha) Z^{l-1}(\theta_b), \mathbf{x})\right) \\ = \alpha f^{l}\left(\theta_a, Z^{l-1}(\theta_a, \mathbf{x})\right) + (1-\alpha) f^{l}\left(\theta_b, Z^{l-1}(\theta_b, \mathbf{x})\right)
    \end{split}
    \end{equation}
    When $f^{l}$ is activation layer such as ReLu, we have
    \begin{equation*}
        \sigma\left(\alpha Z^{l-1}(\theta_a, \mathbf{x}) + (1-\alpha) Z^{l-1}(\theta_b), \mathbf{x})\right) = \alpha \sigma\left(\theta_a, Z^{l-1}(\theta_a, \mathbf{x})\right) + (1-\alpha) \sigma\left(\theta_b, Z^{l-1}(\theta_b, \mathbf{x})\right)
    \end{equation*}
    Here $\sigma(\cdot)$ represent the non-linear function, therefore we have
    \begin{equation*}
        Z^{l}(\theta_b, \mathbf{x}) = f^{l}(\theta_a, Z^{l-1}(\theta_b, \mathbf{x}))
    \end{equation*}
    For layers that perform linear transformation such as linear layer, we can write Eq.~\ref{eq:B_1_2} in matrix form:
    \begin{equation*}
    \begin{split}
        \left(\alpha W^{l}_{\theta_a} + (1-\alpha) W^{l}_{\theta_b}\right) \left(\alpha Z^{l-1}(\theta_a, \mathbf{x}) + (1-\alpha) Z^{l-1}(\theta_b, \mathbf{x})\right) \\ =  \alpha W^{l}_{\theta_a} Z^{l-1}(\theta_a, \mathbf{x}) + (1-\alpha) W^{l}_{\theta_b} Z^{l-1}(\theta_b, \mathbf{x})
    \end{split}
    \end{equation*}
    Then we have
    \begin{equation*}
        W^{l}_{\theta_a} Z^{l-1}(\theta_a, \mathbf{x}) + W^{l}_{\theta_b} Z^{l-1}(\theta_b, \mathbf{x}) = W^{l}_{\theta_a}Z^{l-1}(\theta_b, \mathbf{x}) + W^{l}_{\theta_b} Z^{l-1}(\theta_a, \mathbf{x})
    \end{equation*}
    Derive it, we get
    \begin{equation*}
        \left(W^{l}_{\theta_a} - W^{l}_{\theta_b}\right)\left(Z^{l-1}(\theta_a, \mathbf{x}) -Z^{l-1}(\theta_b, \mathbf{x})\right) = 0
    \end{equation*}
    Therefore, we have
    \begin{equation*}
        Z^{l}(\theta_b, \mathbf{x}) = f^{l}(\theta_a, Z^{l-1}(\theta_b, \mathbf{x}))
    \end{equation*}
    Similarly, when $\theta_a$ and $\pi(\theta_b)$ satisfy LMC, we have
    \begin{equation}
        P^{l}Z^{l}(\theta_b, \mathbf{x}) = Z^{l}(\pi(\theta_b), \mathbf{x}) = f^{l}(\theta_a, Z^{l-1}(\pi(\theta_b), \mathbf{x})) = f^{l}(\theta_a, P^{l-1}Z^{l-1}(\theta_b, \mathbf{x}))
    \end{equation}
    Here $P^{l}$ and $P^{l-1}$ is the permutation matrix defined by $\pi$. Therefore, we could simply define a corresponding natural isomorphism such that each $\tau_{z^{l}}$ is $P^{l}$
\end{proof}

\subsection{Proof for Theorem~\ref{feature_comp}}
\begin{proof}
    According to the partial order between features defined in Sec.~\ref{sec:Complexity},
    when there is more than one natural isomorphisms between $T_{\theta}$ to itself then we have
    \begin{equation}
        \exists l\in [1, L], \exists i, j \in [1, d_l],  s.t. \quad Z^{l}_{i}(\theta) \leq Z^{l}_{j}(\theta), i \neq j.
    \end{equation}
    Next, we need to prove that
    \begin{equation}
        Z^{l}_{i}(\theta) = Z^{l}_{j}(\theta)
    \end{equation}
    Consider the poset $\{Z^{l}_{n}(\theta)| n\in [1, d_l]\}$, since the natural isomorphism defines a one-to-one correspondence between the features, the objects in the same chain are equal. Otherwise, the maximal element $Z^{l}_{max\_S}$ in a chain $S$ must have 
    \begin{equation}
        Z^{l}_{max\_S} \leq Z^{l}_{k},\quad Z^{l}_{k} \notin S 
    \end{equation}
    which contradict to the statement that $Z^{l}_{max\_S}$ is in chain $S$.  Therefore we have
    \begin{equation}
        \exists l\in [1, L], \exists i, j \in [1, d_l],  s.t. \quad Z^{l}_{i}(\theta) = Z^{l}_{j}(\theta), i \neq j.
    \end{equation}
\end{proof}

\section{Experiment Details}
\label{app:exp_detail}
\textbf{Training details for models on CIFAR10}
We train models with the same hyper-parameters. For each model, we train it using SGD with momentum at $0.9$ and weight decay at $1e-4$ for $150$ epochs. The initial learning rate is at $0.1$ and we reduce learning rate at $80$ and $120$ epoch by multiply it with $0.1$. For data augmentation, we only use random horizontal flip with probability set at $50\%$.

Note that the model structures are a little bit different to the structures on ImageNet. For VGGs, the number of layers for the classifier is reduced to $1$ instead of $3$. For ResNet, we apply the conventional change such that the convolution kernel size is set to be $3$ at the first layer.

\textbf{Checkpoint details for models on ImageNet}
For each model, we use pretrained checkpoints on ImageNet provided by pytorch~\citep{TorchVision}. Note that there are two different checkpont for ResNet50 named "ResNet50\_Weights.IMAGENET1K\_V1" and "ResNet50\_Weights.IMAGENET1K\_V2"

\textbf{Details for Iterative Feature Merging}
For IFM, the only hyper-parameter is the $\beta$ in Algorithm~\ref{alg:IFM}. We grid search it in the list [$0.01$, $0.03$, $0.05$, $0.07$, $0.1$, $0.12$, $0.14$, $0.15$, $0.18$, $0.2$].

Note that for ResNet, we only merge the features of the last two block for better merging effect.

\section{Additional Results}
\label{app:add_results}
\subsection{Time Complexity of IFM}
The comutational cost of IFM is relative low since we only need to compute the weight matching distance between features, which is faster than conducting model inference. Table~\ref{tab:time} shows the time consumption of one iteration for different models on CIFAR10 and ImageNet where each result if average over $100$ iteration. The result is conducted with one Geforce RTX 2080Ti.

\begin{table}[htb]
\label{tab:time}
\caption{Time consumption for one iteration with different models on different dataset. The result is averaged over $100$ iterations.}
    \centering
    \begin{tabular}{c|c|c}
    \toprule
      dataset & model & time for one iteration (s) \\
        \hline
       \multirow{4}*{CIFAR10}  & VGG13 & $0.018\pm 0.003$ \\
       & VGG16 & $0.027\pm 0.005$\\
       & ResNet18 & $0.015\pm 0.003$\\
       & ResNet50 & $0.127\pm 0.012$\\
       \hline
       \multirow{4}*{ImageNet} & VGG13 & $0.802\pm 0.053$\\
       & VGG16 & $0.814\pm :0.054$\\
       & VGG19 & $0.822\pm 0.045$\\
       & ResNet18 & $0.015\pm 0.002$\\
       & ResNet50 & $0.127\pm 0.010$\\
    \bottomrule
    \end{tabular}
\end{table}

\subsection{Weight Distance between Features}
\begin{figure}[htb!]
\label{fig:first_layer_distance}
    \centering
\includegraphics[width=0.95\linewidth]{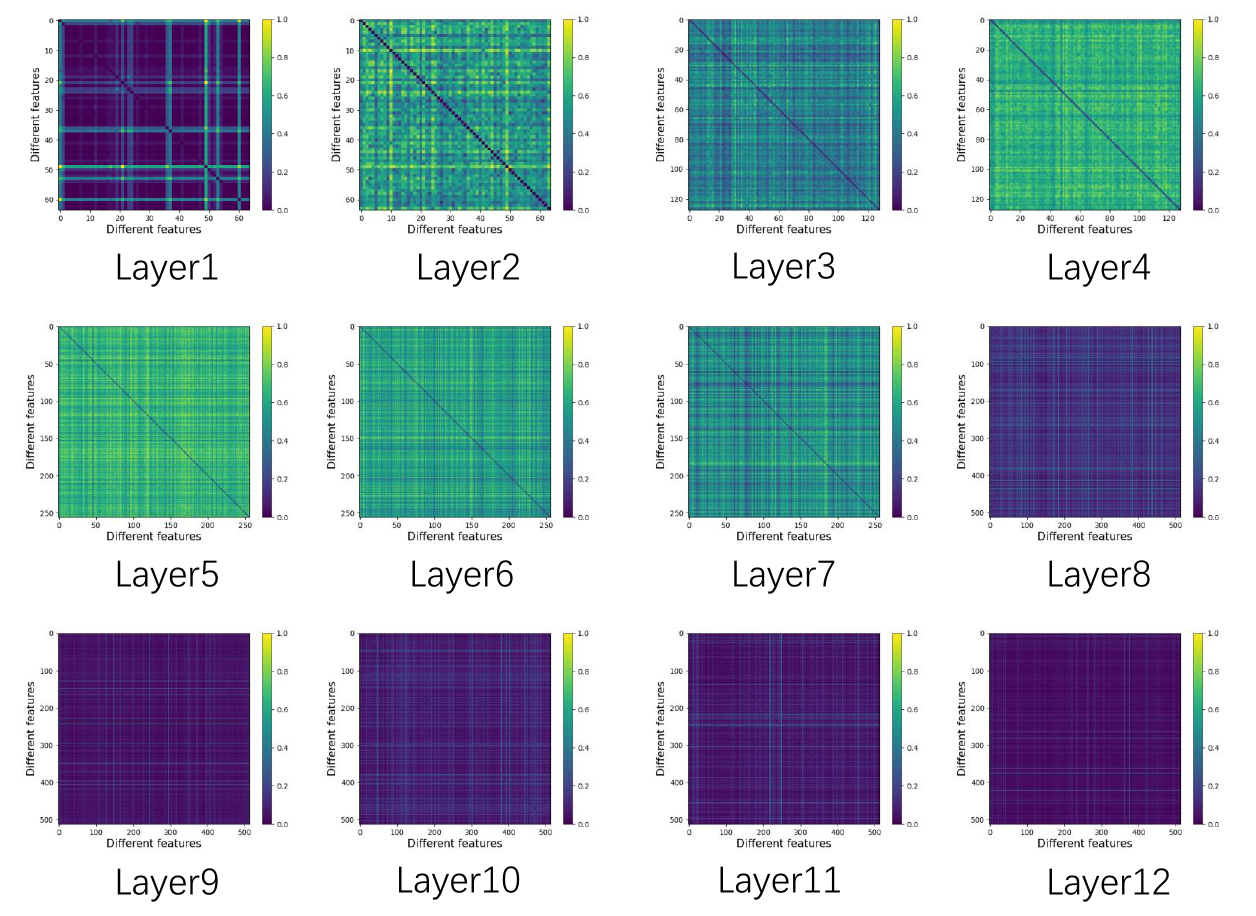}
    \caption{The result of distance weight between features at the first $12$ layers of a VGG16 trained on CIFAR10}
\end{figure}

In Fig.~\ref{fig:existence}, we present the weight distance between features at the last convolution layer of a VGG16 trained on CIFAR10. In this section, we present the result of the first $12$ layers in Fig.~\ref{fig:first_layer_distance}.

\subsection{Guided Backpropagation Results}

\begin{figure}[htb!]
    \centering
    \includegraphics[width=0.95\linewidth]{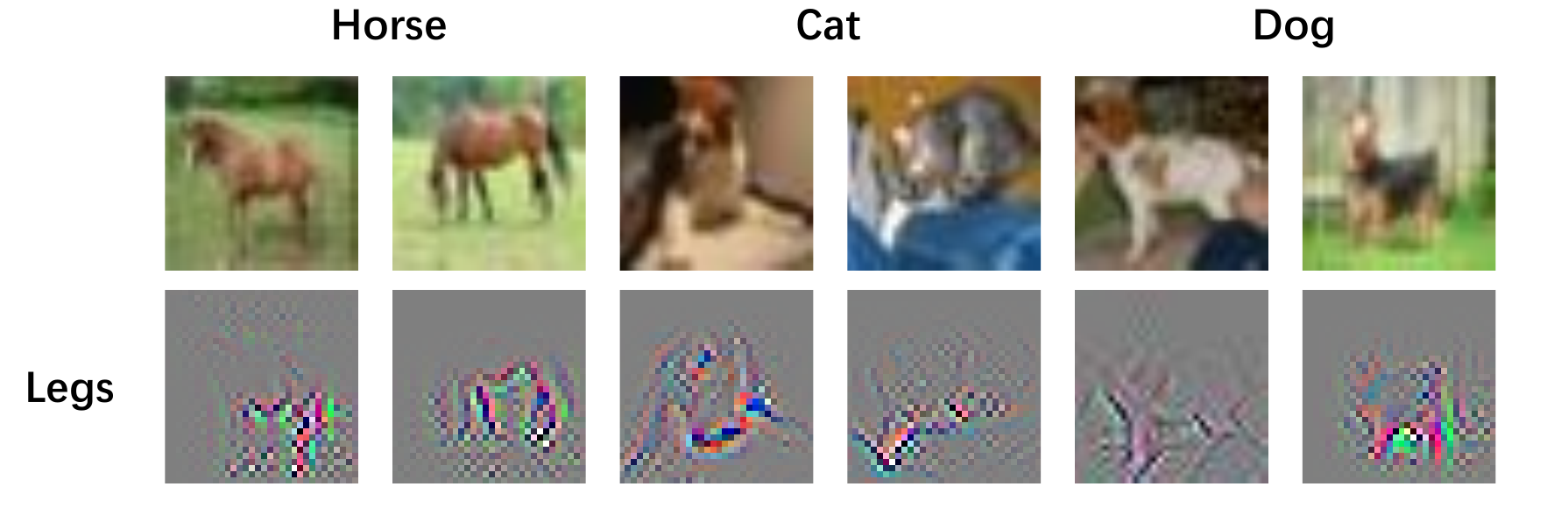}
    \caption{Guided backpropagation result of the VGG16 on CIFAR10. Here we show that a group of functionally equivalent features may corresponds to a semantic across several classes.}
    \label{fig:backprop_v2}
\end{figure}

\begin{figure}[htb!]
    \centering
    \includegraphics[width=0.5\linewidth]{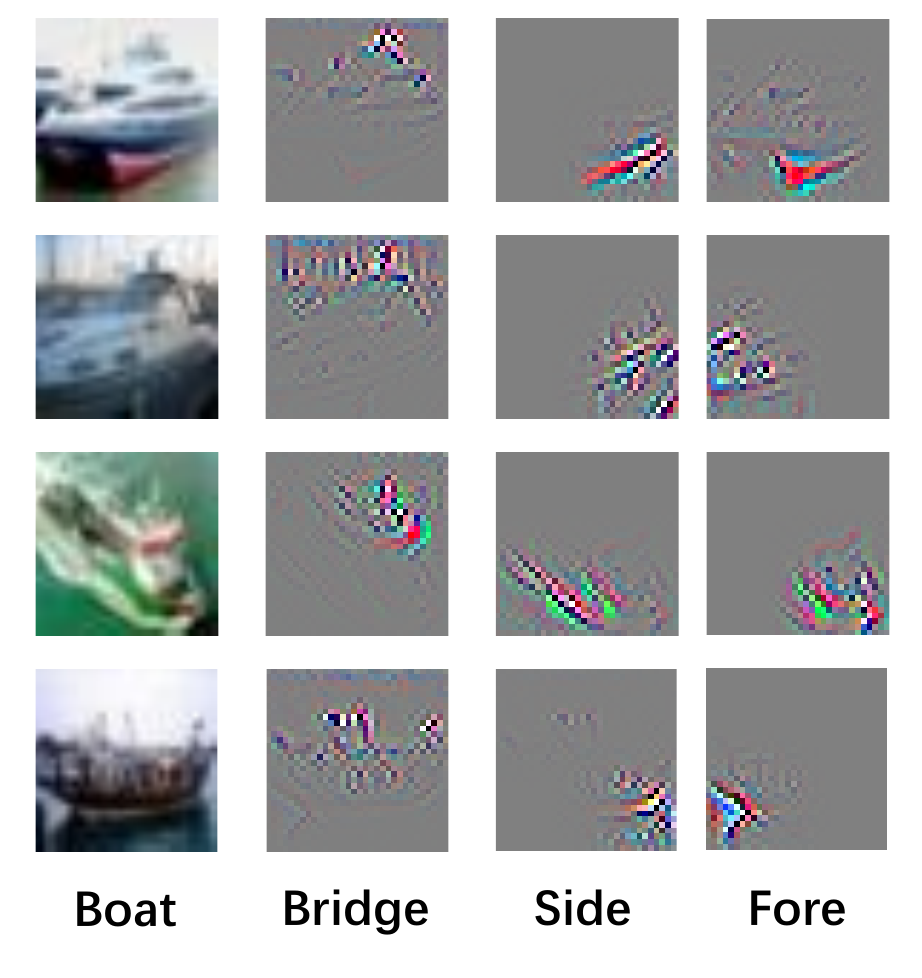}
    \caption{Guided backpropagation result of the VGG16 on CIFAR10. Here we show that a group of functionally equivalent features may corresponds to a semantic regarding boats.}
    \label{fig:backprop_v3}
\end{figure}

In this section, we show more empirical results of the guided backpropagation of different groups of functionally equivalent features. In Fig.~\ref{fig:backprop}, we show that a group of functionally equivalent features may corresponds to a high-level semantic~\citep{guo2018review} in a class. In Fig.~\ref{fig:backprop_v2}, we further show that a group of functionally equivalent features may correspond to a semantic across different classes, where the specific group of features are activated by legs which is presented both on horses, cats and dogs. We also provide results in Fig.~\ref{fig:backprop_v3} that are similar to Fig.~\ref{fig:backprop} where different semantics regarding boats are learned by groups of functionally equivalent features.
Note that all these results are from features at the middle of VGG16, since the result of most features at the last layer cover the whole image. Future works may explore more on the different groups of functionally equivalent features.

\end{document}